\documentclass{article}
\usepackage{arxiv}
\usepackage{geometry}

\usepackage{url} 
\usepackage{lineno}
\usepackage[inline]{trackchanges} 
\usepackage{soul}
\usepackage{titlesec}
\usepackage{graphicx}
\usepackage{graphicx}
\usepackage{titlesec}
\usepackage{makecell}
\titleformat{\section}[block]
{\normalfont\Large\bfseries} 
  {\thesection}{1em}{}         

\titleformat{\subsection}[block]
  {\normalfont\large\bfseries}
  {\thesubsection}{1em}{}

\titleformat{\subsubsection}[block]
  {\normalfont\normalsize\bfseries}
  {\thesubsubsection}{1em}{}
\usepackage{indentfirst,csquotes}

\usepackage{amssymb,amsthm,amsmath}
\usepackage{xcolor,paralist,fancyhdr,etoolbox}

\usepackage[nohyperlinks,nolist]{acronym}
\usepackage{amsfonts,bm,mathtools}
\usepackage{subcaption}
\usepackage{booktabs}
\usepackage{siunitx}
\usepackage{algorithm,algpseudocode}

\usepackage{comment}
\usepackage{float}
\usepackage{hyperref}
\usepackage[capitalize,nameinlink]{cleveref}

\hypersetup{ colorlinks=true, linkcolor=black, filecolor=black, urlcolor=black }

\usepackage{lipsum}

\begin{acronym}
    \acro{KL-DNN}{Karhunen-Loève Deep Neural Network}
    \acro{GCS}{Geological Carbon Storage}
    \acro{IBDP}{Illinois Basin Decatur Project}
    \acro{KL}{Karhunen-Loève}
    \acro{KLE}{Karhunen-Loève Expansion}
    \acro{PDE}{Partial Differential Equations}
    \acro{SVD}{Singular Value Decomposition}
    \acro{DNN}{Deep Neural Network}
    \acro{MSE}{Mean Square Error}
    \acro{RMSE}{Root Mean Square Error}
    \acro{MAE}{Mean Absolute Error}
    \acro{ISGS}{Illinois State Geological Survey}
\end{acronym}

\title{A Trainable-by-Parts Operator Learning Framework: Bridging DeepONet and Karhunen–Loève Expansions for Large-Scale Applications}

\author{
Christian Munoz \\
Civil Engineering Department\
  University of Illinois Urbana-Champaign \\
  Urbana, IL 61801 \\
  \texttt{camunoz4@illinois.edu} \\
   \And
  Alexandre Tartakovsky \\
Civil Engineering Department\
  University of Illinois Urbana-Champaign \\
  Urbana, IL 61801 \\
  Pacific Northwest National Laboratory,\\
  Richland, WA 99352 \\
  \texttt{amt1998@illinois.edu} \\
}

\begin{document}
\maketitle

\begin{abstract}

Training operator-learning models for large-scale problems governed by partial differential equations (PDEs) is challenging due to the curse of dimensionality, memory constraints, and limited training data. These challenges arise in many scientific and engineering applications, including subsurface flow, climate modeling, and geological carbon storage (GCS). In this work, we propose a scalable operator-learning framework based on the Karhunen–Loève Deep Neural Network (KL-DNN) and demonstrate its performance for modeling GCS. The model is trained on a dataset comprising 100 samples of large-scale simulations in a three-dimensional domain with 1.7 million cells and 50 time steps. The KL-DNN method constructs latent spaces using low-rank singular value decomposition of static properties and a nested Karhunen–Loève expansion for dynamic pressure fields, enabling full-resolution predictions without subsampling or spatial coarsening. The KL-DNN model achieves an average \ac{RMSE} of 1.1 psi for pressure ($\approx 0.04$\% relative error with respect to the average pressure in the domain) and RMSE of 0.0146 for CO$_2$ saturation ($\approx$ 5\% relative error with respect to the average saturation inside the plume). The model requires $\approx 20$ minutes of training on a single GPU, representing a 19\% reduction in the pressure errors, 7\% reduction in the saturation error, and a two-order-of-magnitude speedup compared to DeepONet trained on the same dataset. These results, along with inference time of less than one minute, establish the proposed model as a practical and accurate solution for large-scale PDE problems, enabling rapid uncertainty quantification, history matching, and real-time decision support.

\end{abstract}
\maketitle

\section*{Plain Language Summary}

We propose a faster, more scalable ML model for complex physical systems, such as underground carbon storage. Traditional physics-based simulations for these systems are accurate but slow and computationally expensive, making them impractical when many repeated simulations are needed (for example, to assess uncertainty or test different scenarios).
We introduce a framework called a Karhunen-Loève Deep Neural Network (KL-DNN). The key idea is to compress very large, high-dimensional data (such as 3D pressure and CO$_2$ saturation fields evolving over time) into a much smaller set of variables that still capture the essential patterns. The model is trained in parts, allowing the most computationally demanding steps to be handled separately and efficiently.
The method is tested on a realistic, large-scale carbon storage problem with millions of grid cells. Despite using only 100 training simulations, the model accurately predicts pressure and CO$_2$ distribution across space and time. It achieves lower prediction errors than an established approach (DeepONet) while reducing training time from days to about 20 minutes and producing results in under a minute.
Overall, this framework enables rapid, reliable predictions for large-scale geophysical systems, supporting applications such as uncertainty quantification, model calibration, and real-time decision-making.

\section{Introduction}
\acresetall
\acused{GCS}
\acused{KL}
\acused{IBDP}
\acused{KLE}
\acused{KL-DNN}
\acused{RMSE}
Physics-based simulations of large-scale problems, while accurate, are computationally expensive. The main reason is that the computational cost of numerical solutions to partial differential equations (PDEs) increases exponentially with the number of cells in the domain discretization, a phenomenon known as the curse of dimensionality.

In this work, we focus on problems that require millions of nodes and tens of time intervals to adequately represent the system's properties and states. Geological Carbon Storage (GCS) is one example of such problems. In \ac{GCS} applications, the system needs a fine discretization in critical areas, such as injection zones and geological faults, which often results in a large number of nodes in numerical models.
High heterogeneity of subsurface parameters (such as porosity and permeability) and the inherent uncertainty in their distribution make subsurface modeling especially challenging \cite{sudicckyillman2010,woodburyulrych2000,hess-27-255-2023}. The practical application of subsurface models requires multiple model evaluations across different parameter combinations. For example, numerous model runs are needed to estimate uncertainty in model predictions, investigate different scenarios, and perform history matching.    

The need to perform multiple simulations and their high computational cost motivate the development of operator-learning models that can predict system states (e.g., CO$_2$ pressure and saturation in GCS) as functions of system parameters (e.g., reservoir porosity and permeability). Several operator learning models have been proposed for GCS, including Fourier Neural Operator \cite{TANG2024130641}, convolutional encoder-decoder networks \cite{MO2018}, and DeepONet \cite{KADEETHUM2024213007}. 
Among these approaches, only the DeepONet model in \cite{KADEETHUM2024213007} addresses problems at the scale considered in this work.
Specifically, in \cite{KADEETHUM2024213007}, the DeepONet model was trained for the Illinois Basin Decatur Project (\ac{IBDP}) CO$_2$ injection site using a training dataset consisting of numerical simulations performed on a three-dimensional domain discretized into 1.7 million cells, with solutions reported at 50 time steps.

DeepONet performs operator learning in a reduced (latent) parameter space, determined by the branch networks, and in a reduced state-variable space, determined by the trunk network. 
Other related latent-space operator learning models for parametric PDEs include PCA-Net \cite{bhattacharya2021model} and KL-DNN \cite{WANG2024117147,WANG2025105024}. These two methods use closely related PCA and \ac{KL} expansions \cite{gerbrands1981relationships} to find latent spaces. 
In DeepONet, PCA-Net, and KL-DNN, operator learning is performed in a latent space using a fully connected feed-forward deep neural network (DNN), which requires only a few fully connected layers. However, learning a latent representation for large-scale problems can require prohibitively large computer memory and training time.

Training DeepONet for large-scale problems is particularly challenging because the latent representations of states and parameters are jointly learned and combined with DNN training. In PCA-Net and KL-DNN, the latent representations are learned separately for states and parameters via singular value decomposition (SVD). However, performing SVD on large datasets may require substantial memory that may not be available even on large supercomputers. 
   
DeepONet has been applied to large-scale problems using several strategies. DeepONet with domain decomposition (DD-DeepONet) splits large domains into subdomains, reducing memory requirements and training time \cite{yang2025dd}. Time-integration DeepONet variants (TI-DeepONet) focus on derivative operators to avoid long sequential rollouts \cite{nayak2025ti}. Ensemble and mixture-of-experts architectures further localize learning for heterogeneous domains \cite{sharma2025}. Separable DeepONet (SepONet) factorizes spatial dimensions to lower complexity and memory footprint \cite{yu2024separable}. In \cite{KADEETHUM2024213007}, a graph-based topology embedding was introduced in DeepONet to capture local and global interactions. A dynamic resampling of training subsets was used to manage the memory required for DeepONet training and the associated computational costs when handling domains
with millions of cells and multiple time steps. 

In PCA-Net, a localized PCA on spatial patches was used to reduce computational cost and memory requirements \cite{dhingra2025localized}. This patch-based approach achieved up to four times faster training than global PCA models and improved accuracy near critical regions by preserving local variability. 

In this study, we propose a general trainable-by-parts operator learning framework for large-scale PDE problems. As the basis of this framework, we adapt the KL-DNN/PCA-Net method. In large-scale applications, the KL-DNN and PCA-Net methods are nearly identical, as both use SVD to construct PCA or KL representations of parameter and state fields. Furthermore, we demonstrate that the resulting model is a special case of DeepONet and can be considered a generalized operator-learning model. 

The main innovations of this work are the following: 
\begin{itemize}
\item We propose a two-step approach for performing SVD decomposition on large-scale problems. In the first step, the multi-dimensional KLE is decomposed into the lower-dimensional KLEs. In the second step, we estimate eigenvalues and eigenfunctions in the low-dimensional KLEs using a low-rank SVD. This approach enables the SVD decomposition of large datasets without significantly compromising accuracy, a task that is often infeasible even on supercomputers due to memory limitations.
\item We establish a correspondence between the PCA-Net/KL-DNN and DeepONet models. This allows for separately training the trunk and branch of DeepONet and extending it to large-scale problems. 
\end{itemize}

We demonstrate the effectiveness of the proposed model by comparing its accuracy and training time to those of the DeepONet model trained on the IBDP dataset, which consists of 100 samples, each sample computed on a 1.7 million-node-mesh and 50 time steps \cite{KADEETHUM2024213007}. We demonstrate that our model requires 2 orders of magnitude less training time and is up to 19\% more accurate than the DeepONet trained on this large dataset using dynamic resampling.

This work is organized as follows: Section \ref{sec:dynamic_kl_dnn} presents the scalable operator learning model. Section \ref{sec:ibdp_dataset} describes training and the predictions of the operator learning model using the \ac{IBDP} dataset. Conclusions are given in Section \ref{sec:conclusion}.

\section{Operator learning model}\label{sec:dynamic_kl_dnn}

We present the operator learning method for the general \ac{PDE} problem,
\begin{equation}\label{eq:PDE}
L[u(\bm{x},t),y_1(\bm{x}),...,y_{N_p}(\bm{x})] = 0, 
\end{equation}
subject to appropriate initial and boundary conditions. Here, $L$ is the differential operator, $u(\bm{x},t)$ is the state variable, and $y_1(\bm{x}),...,y_{N_p}(\bm{x})$ are the $N_p$ space-dependent parameter fields.

We assume that a labeled dataset $D=\{ \bm{y}_1^{(i)},...,\bm{y}^{(i)}_{N_p}  \rightarrow \bm{u}^{(i)}  \}_{i=1}^{N_{train}}$ exists consisting of $N_{train}$ samples of the numerical PDE solution obtained on a $N_x \times N_y \times N_z \times N_t$ space-time mesh.   

Our goal is to develop an operator learning model $\mathcal{G}_u [\bm{y}_1,...,\bm{y}_{N_p}] \left (\bm{x}, t \right )$ for large-scale problems where $N_x$, $N_y$, $N_z$, and $N_t$  may be on the order of millions. Below, we discuss the challenges of training operator learning models for large-scale problems and how the proposed model addresses them.  

To motivate the proposed approach, we first review the KL-DNN formulation of a neural operator $\mathcal{G}_u [y(\bm{x})](\bm{x}, t)$  for $N_p=1$ (a single parameter field). In this method, $u$ and $y$ are treated as realizations of random processes $U$ and $Y$, and $\bm{y}^{(i)}$ and $\bm{u}^{(i)}$ in the training dataset $D$ are generated as independent samples of $Y$ and $U$, respectively. The operator $\mathcal{G}_u [y](\bm{x}, t)$ is given by the space-time KLE of $U$  \cite{TARTAKOVSKY2024112723}:
\begin{equation}\label{eq:truncation_error}
\mathcal{G}_u [y](\bm{x}, t) = \bar{u}(\bm{x},t) + \sum^{N_\eta}_{i = 1} \eta_i[\bm\xi (y);\bm\theta]\phi_i(\bm{x},t) \sqrt{\lambda_i},
\end{equation}
where $\bar{u}(\bm{x},t)$ is the ensemble mean of $U$, $\phi_i(\bm{x},t)$, ${\lambda_i}$ are the $N_\eta$ leading eigenfunctions and eigenvalues of the covariance of $U$, and $\eta_i$ are the KLE coefficients. The eigenfunctions and eigenvalues are found from the eigenvalue decomposition of the sample covariance matrix of $U$ or the SVD of $\bm{u}^{(i)}$ samples.  Therefore, the maximum number of eigenpairs that can be computed is $N_{train}$, i.e., $N_\eta \le N_{train}$.
For a sufficiently large $N_{train}$, $N_\eta$ can be determined from the Mercer theorem \cite{mercer1909functions} according to the desired tolerance $rtol$ as
\begin{equation}\label{eq:tol}
rtol = {\sum_{i=N_\eta+1}^{\infty}\lambda_{i} }/{\sum_{i=1}^{\infty}\lambda_{i}}.
\end{equation}
However, for large-scale problems, $N_{train}$ is relatively low due to the high computational cost of generating samples. Therefore, $N_{train}$ is usually a limiting factor in selecting $N_\eta$. Specific constraints for selecting $N_\eta$ will be discussed later in this section.

\begin{figure}[htbp]
\centering
\includegraphics[width=0.8\textwidth]{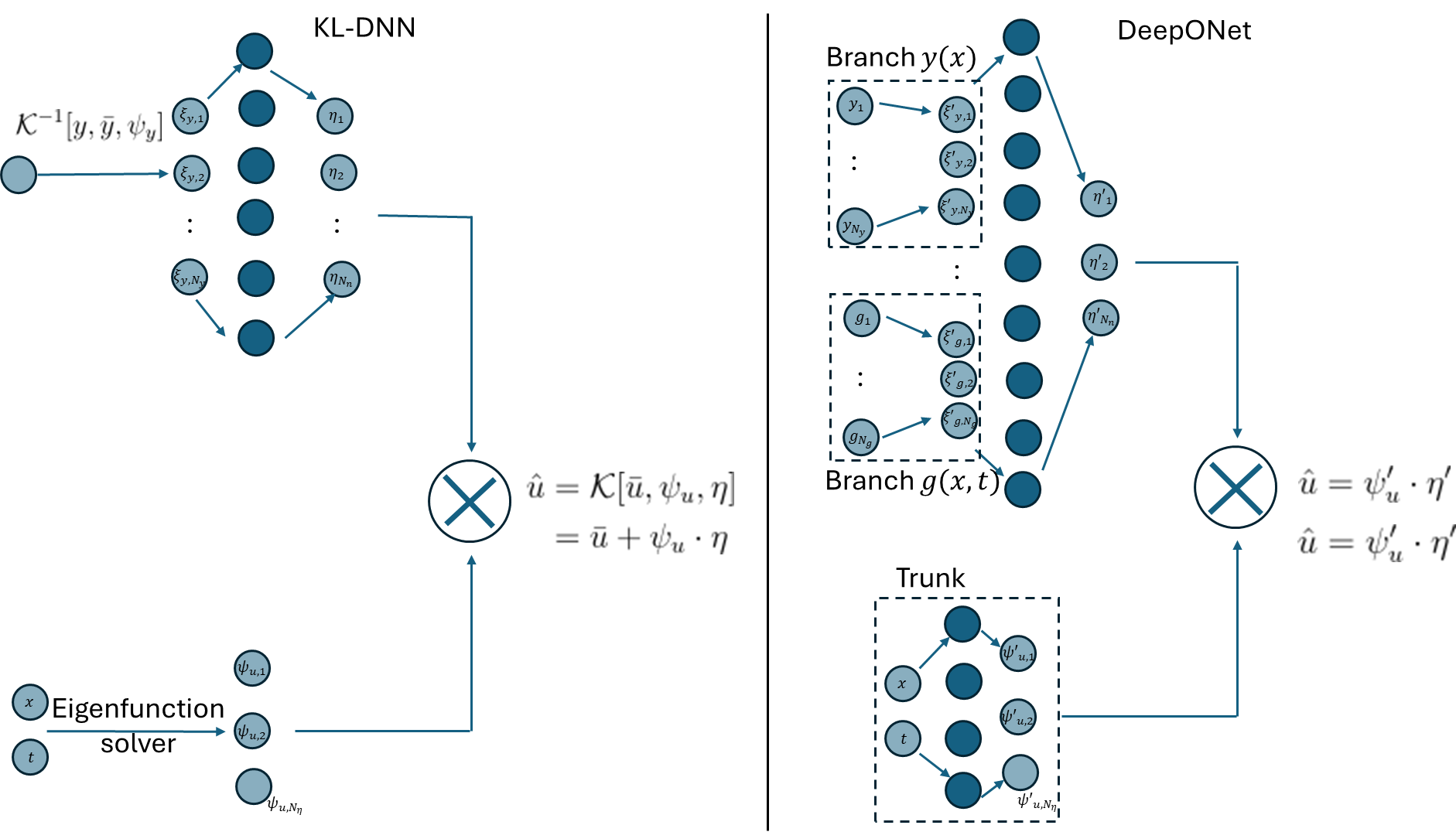}
\caption{The KL-DNN operator-learning framework (left) and the vanilla DeepONet architecture (right). In KL-DNN, the KLE of the state variable replaces the DeepONet trunk. The inverse KLE of the input variable $y$ in KL-DNN replaces the encoder part of the DeepONet branch. The DNN part of KL-DNN is analogous to the fully connected layers in the DeepONet branch.}
\label{fig:kl_dnn_framework}
\end{figure}

The KLE coefficients $\bm\eta = [\eta_1, ..., \eta_N]^T$ are mapped to the vector of coefficients $\bm{\xi} = (\xi_1, ..., \xi_{N_\xi})^T$ in the truncated KLE of the parameter field $y$, 
\begin{equation}
  \label{eq:tdkle}
 y(\bm{x})  \approx \mathcal{K}_y\left(\bm{x}, \bm{\xi} \right ) = \bar{y}(\bm{x}) + \sum^{N_\xi}_{i = 1} \xi_i \chi_i(\bm{x}) \sqrt{\beta_i} ,
\end{equation}
using the fully connected feed-forward DNN $\bm{\eta} = \mathcal{NN}(\bm\xi;\bm\theta)$
with the output vector  $\bm\eta$ and the parameters $\bm\theta$.   
In the \ac{KLE} of $y$, $\bar{y}(\bm{x})$ is the mean of $Y$, and  $\chi_i(\bm{x})$, and ${\beta_i}$ are the eigenfunctions and eigenvalues of $C_y(\bm{x},\bm{x}')$, the (known) covariance function of $Y$. The KL-DNN model is shown in Figure \ref{fig:kl_dnn_framework}.

Similar to other neural operator models, including PCA-Net and DeepONet, KL-DNN can be trained using  the full dataset $D$ by solving the minimization problem
\begin{equation}\label{eq:DNNloss}
\boldsymbol\theta^* = 
\min_{\boldsymbol\theta} \sum_{i=1}^{N_\text{train}} 
||u^{(i)}(\bm{x},t) - 
 \mathcal{G}_u [\bm{y}^{(i)}](\bm{x}, t;\bm\theta)
||^2_2 + \frac1{\sigma^2}||\bm\theta||_2^2,
\end{equation}
where the $\ell_2$ norm of any function $c(x,t)$ is defined as  
$
    || c(x,t)||_2 =\sqrt{\int_0^T \int_\Omega (c(x,t) )^2 dx dt}
$ 
and $1/\sigma^2$ is a coefficient in the regularization term.   
The advantage of having pre-computed KLEs with the orthogonal eigenfunctions is that this expensive-to-compute function norm can be replaced with a much simpler $\ell_2$ vector norm as
\begin{equation}\label{eq:norm_reduction}
||u^{(i)}(\bm{x},t) - 
 \mathcal{G}_u [\bm{y}^{(i)}](\bm{x}, t;\bm\theta)
||^2_2
=
[\bm\eta^{(i)}-\hat{\bm\eta}(\bm\xi^{(i)};\bm\theta)]^T \bm\Lambda [\bm\eta^{(i)}-\hat{\bm\eta}(\bm\xi^{(i)};\bm\theta)] = 
||
[\bm\eta^{(i)}-\hat{\bm\eta}(\bm\xi^{(i)};\bm\theta)
||^2_{\bm\Lambda},
\end{equation}
where $\bm\Lambda$ is the diagonal matrix with the diagonal elements $\Lambda_{ii} = \lambda_i$ and $\bm\eta^{(i)}$ is the vector of coefficients in the KLE of $\bm{u}^{(i)}$ \cite{reyna2025}.  It follows from Eq \eqref{eq:norm_reduction} that the parameters $\bm\theta$ in the KL-DNN model can be found using the reduced (latent)-space dataset $D_l=\{\bm\xi^{(i)} \rightarrow \bm\eta^{(i)}  \}_{i=1}^{N_{\text{train}}}$ by solving the minimization problem
\begin{align}\label{eq:surrogate_loss}
L(\bm\theta) = \sum_{i=1}^{N_{\text{train}}} || \mathcal{NN}(\bm{\xi}^{(i)};\bm\theta) - \boldsymbol{\eta}^{(i)} ||^2_{\bm\Lambda} + \frac1{\sigma^2}||\bm\theta||_2^2. 
\end{align}
The reduced dataset is obtained by applying the inverse KL operator $\mathcal{K}^{-1}$ to the elements of the $D$ dataset: $\bm{\xi}^{(i)} = \mathcal{K}^{-1}[\bm{y}^{(i)}]$ and $\bm{\eta}^{(i)} = \mathcal{K}^{-1}[\bm{u}^{(i)}]$. For example, for the KLE of $u$,
\begin{equation}
  \label{eq:kleu}
 u(\bm{x},t)  \approx \mathcal{K}_u\left(\bm{x}, t,\bm{\eta} \right ) = \bar{u}(\bm{x},t) + \sum^{N_\eta}_{i = 1} \eta_i \phi_i(\bm{x},t) \sqrt{\lambda_i} ,
\end{equation}
the inverse operator acting on $\bm{u}^{(i)}$ is defined as
 \begin{align}
  \mathcal{K}^{-1} \left[ \bm{u}^{(i)} \right] \coloneq \arg\min_{\bm{\eta} } \left( ||  \bm{u}(\bm{x}, t)^{(i)} - \mathcal{K}_u \left[ \bm{x},t,\bm{\eta}\right]  ||_2^2 + \gamma || \bm\eta||_2^2  \right),
   \label{eq:inverseKL}
\end{align}
where $\gamma$ is a regularization coefficient. Using the loss function in Eq \eqref{eq:surrogate_loss} instead of that in Eq \eqref{eq:DNNloss} significantly reduces the training time and the memory requirements. 

The mean and covariance of $U$  can be computed by replacing $y$ and $u$ in the PDE problem \eqref{eq:PDE} with their stochastic counterparts, and solving the resulting stochastic PDE for $\overline{u}$ and $C_u$ using methods such as Polynomial Chaos \cite{xiu2003modeling}, Moment Equations \cite{TARTAKOVSKY2003182}, or Monte Carlo \cite{Tartakovsky2020JCP_PICKLE}. Most commonly, the Monte Carlo method is employed where the dataset $D$ is used to approximate $\overline{u}$ and $C_u$ as the sample mean vector $\bar{\bm{u}}  \in \mathbb{R}^{ N_{xyzt} }$,
\begin{equation}
 \bar{\bm{u}} = \frac{1}{N_\text{train}}\sum_{i=1}^{N_\text{train}}   \bm{u}^{(i)},
\end{equation}
and the sample covariance matrix $\mathbf{C}_u  \in \mathbb{R}^{ N_{xyzt} \times N_{xyzt} }$,
\begin{equation}
    \mathbf{C}_u = \frac1{N_{train} - 1} \mathbf{X}^T \mathbf{X},
\end{equation}
where $N_{xyzt} = N_x \times N_y \times N_z \times N_t$ and the centered data matrix $\mathbf{X} \in \mathbb{R}^{N_{train} \times N_{xyzt}}$ consists of elements $\{ X_{ij} = u^{(i)}_j - \overline{u}_j \}_{i=1,j=1}^{N_{train},N_{xyzt}}$. 

The eigenvalues ${\lambda_i}$ and eigenfunctions $\phi_i(\bm{x},t)$ can be found from the eigenvalue problem
\begin{equation}\label{eq:cev_problem}
  \int_D \int_T C_u(\bm{x}, \bm{x}', t , t') \phi (\bm{x}',t') \, \mathrm{d} \bm{x}'  \mathrm{d} t'= \lambda \phi(\bm{x},t).
\end{equation}
The numerical solution of Eq \eqref{eq:cev_problem}  can be obtained by discretizing it as 
\begin{equation}\label{eq:cev_discrete}
  (\bm{W}^{1/2} \times \bm{C}_u \times \bm{W}^{1/2}) \times (\bm{W}^{1/2} \times \bm{\phi}) = \lambda ( \bm{W}^{1/2} \times \bm{\phi}),
\end{equation}
where $\bm{\phi}$ is the eigenvector, $\bm{W}$ is a diagonal matrix with the diagonal elements $W_{jj} = w_j$, and $w_j$ is the volume of $j$th cell in the numerical solutions $\bm{u}^{(i)}$ \cite{safta2026numerical}. Here, we assume that all solutions $\bm{u}^{(i)}$ are obtained on the same numerical mesh. For large-scale applications, the $\mathbf{C}_u$ size makes solving this eigenvalue problem impractical. 

The common solution to reducing the computational cost of solving the eigenvalue problem is to use the SVD decomposition:
$$\tilde{\bm{X}} = \bm{U}\bm{S} \bm{V}^T$$
where $\tilde{\bm{X}} = \bm{W}^{1/2} \bm{X}$, $\bm{U} \in \mathbb{R}^{ N_{xyzt} \times N_{xyzt} }$ is the orthogonal matrix, $\bm{S} \in \mathbb{R}^{N_{train} \times N_{xyzt}}$ is the diagonal matrix containing singular values $s_i$ (in the descending order), and $\bm{V} \in  \mathbb{R}^{ N_{train} \times N_{train} }$ is the orthogonal matrix containing eigenvectors. 
The eigenvalues are found as
\begin{equation}
    \lambda_i = \frac{s_i^2}{N_{train} - 1}.
\end{equation}
The mean and eigenfunctions $\overline{u}(\bm{x},t)$ and $\phi_i(\bm{x},t)$ are approximated via a polynomial interpolation of their nodal values given by $\overline{\bm{u}}$ and the eigenvectors, respectively. It should be noted that in the standard PCA and SVD-based expansions, $\bm{W}$ is replaced with the identity matrix. The resulting eigenvalues and eigenvectors are, in general, different from those given by the solution of the eigenvalue problem \eqref{eq:cev_discrete}. 

While being faster than eigenvalue solvers, the SVD decomposition still requires storing the $N_{xyzt} \times N_{xyzt}$ matrix $\bm{U}$, i.e., it does not address memory limitations. Also, to avoid overfitting in $\mathcal{NN}(\bm\xi)$ training, $N_\eta$ must be kept smaller (often by an order of magnitude) than $N_{train}$, in turn requiring only $N_{train}$ largest eigenvalues and the corresponding eigenfunctions. This allows reducing both the computation time and the memory requirements by replacing the (full) SVD decomposition with the low-rank SVD decomposition \cite{brand2006fast}
$$\tilde{\bm{X}} = \bm{U}_r\bm{S}_r \bm{V}_r^T,$$
where $\bm{U}_r \in \mathbb{R}^{ N_{xyzt} \times N_\eta }$ contains the first $N_\eta$ columns of $\bm{U}$, $\bm{S}_r \in \mathbb{R}^{N_{\eta} \times N_{\eta}}$ is the diagonal matrix containing first $N_\eta$ singular values, and $\bm{V}_r \in  \mathbb{R}^{ N_{train} \times N_{\eta} }$ is the orthogonal matrix containing eigenvectors. The main memory saving here is achieved by replacing $\bm{U} \in \mathbb{R}^{ N_{xyzt} \times N_{xyzt} }$ with $\bm{U}_r \in \mathbb{R}^{ N_{xyzt} \times N_\eta }$. 

However, for $N_{xyzt}$ on the order of tens of millions, performing the low-rank SVD could require prohibitively large computer memory. 
To further reduce the computational cost of constructing KLEs, we decompose multi-dimensional KLEs into lower-dimensional KLEs \cite{ZHENG2017221}. For example, the KLE of the stochastic process $U$  
\begin{equation}\label{eq:KLE_general}
U(\bm{x}, t) = \bar{u}(\bm{x},t) + \sum^{N_\eta}_{i = 1} \tilde{\eta_i}\phi_i(\bm{x},t) \sqrt{\lambda_i},
\end{equation}
where $\{ \tilde{\eta_i} \}_{i=1}^{N_\eta}$ are uncorrelated random variables, can be approximated as 
\begin{equation}\label{eq:nested_KLE}
U(\bm{x}, t) = \bar{u}(\bm{x},t) + \sum^{N_\eta}_{i = 1} {\tilde{\eta}'_i}(t) \phi'_i(\bm{x},t) \sqrt{\lambda'_i(t)}.
\end{equation}
The random processes $\{ \tilde{\eta_i} \}_{i=1}^{N_\eta}$ are further decomposed using one-dimensional KLEs:
\begin{equation}\label{eq:extra_KLE}
    \tilde{\eta}'_i(t) \approx \bar{\eta_i} + \sum_{j=1}^{N_\gamma} \gamma_{ij} \psi_{ij}(t) \sqrt{\delta_{ij}}, 
\end{equation}
where $\bar{\eta_i}(t)$ is the mean of $\tilde{\eta}'_i(t)$ and $\psi_{ij}(t)$ and $\delta_{ij}$ are the eigenfunctions and eigenvalues of the covariance $C_{\tilde{\eta}_i}(t,t')$ of $\tilde{\eta}'_i(t)$. We note that the ensemble mean of $\tilde{\eta}'_i(t)$ is zero, but its sample approximation might not be. 

In Eq \eqref{eq:nested_KLE}, $\phi'_i(\bm{x},t)$ and $\lambda'_i(t)$ are given by the solution of the eigenvalue problem 
\begin{equation}
  \int_D  C_u(\bm{x}, \bm{x}', t,t ) \phi' (\bm{x}',t) \, \mathrm{d} \bm{x}' = \lambda'(t) \phi'(\bm{x},t).
\end{equation}
The dimensionality of this eigenvalue problem is $d$ (the number of spatial dimensions) while the dimensionality of the eigenvalue problem in Eq \eqref{eq:cev_problem} is $d+1$. 
For the KLE in Eq \eqref{eq:nested_KLE}, the largest matrix in the low-rank SVD has size $N_{xyz} \times N_\eta $, where $N_{xyz} = N_x\times N_y \times N_z$. This matrix is $N_t$ times smaller than the corresponding matrix in the full space–time KLE.
There is an extra cost of computing the eigenpairs of the KLE \eqref{eq:extra_KLE}; however, since this KLE is one-dimensional, this extra cost is not significant. 

We first compute $\{ \phi'_i(\bm{x},t_j), \lambda'_i(t_j) \}_{j=1}^{N_t}$ by performing low-rank SVD of the centered data matrix 
$\mathbf{X}_i \in \mathbb{R}^{N_{train} \times N_{xyz}}$ ($i=1,...,N_t$) 
consisting of elements $\{ X_{i,kl} = u_{i,kl} - \overline{u}_{i,l} \}_{k=1,l=1}^{N_{train},N_{xyz}}$ for each timestep $t_i$, where $u_{i,kl}$ is the element of the $k$th solution sample at $l$th spatial element at time $t_i$ and $\overline{u}_{i,l}$ is the sample mean at $l$th spacial element at time $t_i$.
Next, $\{ \tilde{\eta}^{'(k)}_i(t_j)  \}_{i=1,j=1,k=1}^{N_\eta,N_t,N_{train}}$ are computed by the 
least-squares fitting of KLE \eqref{eq:nested_KLE} to $\bm{u}^{(i)}$ as 
\begin{equation}
\tilde{\eta}^{'(k)}_1(t_j),...,\tilde{\eta}^{'(k)}_{N_\eta} (t_j) =
\min_{\tilde{\eta}'_1,...,\tilde{\eta}'_{N_\eta}} 
\sum_{l=1}^{N_{xyz}} \left[ u^{(k)}_{j,l} - 
\bar{u}(\bm{x}_l,t_j) -
\sum^{N_\eta}_{i = 1} {\tilde{\eta}'_i}
\phi'_i(\bm{x}_l,t_j) \sqrt{\lambda'_i(t_j)} \right]^2.
\end{equation}
 
Finally, $\psi_{ij}$ and $\delta_{ij}$ are computed using the low-rank SVD of the data matrix 
$\bm{X}_k \in \mathbb{R}^{N_{train} \times N_t} $ ($k=1,...,N_\eta$) 
consisting of elements $\{ X_{k,ij} = \eta^{(i)}_{k,j} - \overline{\eta}_{k,j} \}_{i=1,j=1}^{N_{train},N_t}$ for each $\eta_k$ component.

The final form of the neural operator is  
\begin{equation} \label{eq:nested_u}
    \mathcal{G}_u [y](\bm{x}, t)  = \bar{u}(\bm{x}, t) + \sum_{i=1}^{N_\eta} \left[ \bar{\eta_i} + \sum_{j=1}^{N_\gamma} \gamma_{ij}[\bm\xi(y)] \sqrt{\delta_{ij}} \psi_{ij}(t) \right] \sqrt{\lambda_i(t)} \phi_i(\bm{x}, t).
\end{equation}

Next, we discuss the criteria for selecting the total number of terms in the KLEs of $u$ and $y$, i.e.,   $N_\xi$ and $N_\eta$ in the full-dimensional KLEs \eqref{eq:kleu} and \eqref{eq:tdkle} or $N_\eta \times N_\gamma$ in the low-dimensional KLE \eqref{eq:nested_u}. We stated earlier that $N_\xi$ and $N_\eta$ must be less than or equal to $N_{train}$.  On the other hand, $N_\xi$ and $N_\eta$ are the dimensions of the input and output layers of the DNN, which is trained with $N_{train}$ samples. The rule of thumb for selecting the DNN size as a function of the number of samples in the dataset is to keep the total number of trainable parameters in the DNN lower than $N_{train}$ by at least a factor of 10 \cite{alwosheel2018your}, which requires $N_\eta + N_\xi << N_{train}$. In practice, this rule of thumb is too restrictive.
The values of $N_\eta$ and $N_\xi$ are treated as hyperparameters and are chosen to minimize the error in predicting unseen solutions, the so-called testing error.

At this point, it is appropriate to discuss the relationship between the KL-DNN model and the PCA-Net and DeepONet models. The PCA-Net and KL-DNN methods are structurally similar and use related PCA and KL-DNN expansions. Moreover, both methods rely on SVD to construct the expansions. The only differences between the PCA-Net and KL-DNN models are how the data is generated (in the KL-DNN model, Monte Carlo sampling is used, whereas there are no such restrictions in the PCA-Net) and how the data matrix $\bm{X}$ is constructed. In KL-DNN, the elements of $\bm{X}$ are weighted with the size of numerical elements for the singular values and eigenvectors to approximate the solution of the eigenvalue problem, while in PCA-Net, no weights are applied to the data. 

The vanilla DeepONet method is shown in Figure \ref{fig:kl_dnn_framework}. In both the KL-DNN and DeepONet methods, the operator is given as the product of the parameter and basis function vectors. In the DeepONet, the basis functions are the output of the trunk DNN, and the parameter vector $\bm\eta'$ is the output of the branch DNN. In KL-DNN, the basis functions are obtained by SVD, and the vector $\bm\eta$ is obtained by applying the inverse KL operators to the state variables.    
 Despite the similarity, KL-DNN has a computational advantage over DeepONet.
The ``training'' of all components in the KL-DNN model (solving eigenvalue problems, applying inverse KL operations, and the DNN training) is done independently and without accessing the entire dataset $D_{\text{train}}$. In DeepONet, the trunk and branch DNNs (including all sub-branch DNNs) must be trained jointly using an expensive-to-evaluate loss function similar to \eqref{eq:DNNloss}, which requires access to the entire $D_\text{train}$ dataset and, in turn, large computer memory. 
 
Therefore, the KL-DNN architecture in Figure \ref{fig:kl_dnn_framework} can be considered a variant of DeepONet that enables separate training of the trunk, branch encoder, and branch DNN.

Finally, we discuss the KL-DNN model for $N_p>1$, i.e., a problem with more than one parameter field. If the parameter fields are uncorrelated, then each parameter field $y_i$ should be represented with a separate KLE with the KL coefficient vector $\bm\xi_i$, yielding the vector containing KL coefficients of all parameters, $\bm\xi_T = [\bm\xi_1,...,\bm\xi_{N_p}]$. The KL-DNN model is then modified by using $\bm\xi_T$ as input to the DNN, i.e., $\bm\eta = \mathcal{NN}(\bm\xi_T)$.  It should be noted that the DNN size depends on the input layer size, and training a DNN with a larger input layer requires more samples. Therefore, there is a need to reduce the size of the latent space representing all parameter fields (and the input layer of the DNN model), especially for large-scale problems. 

The correlated parameter fields can be expressed with a single KLE \cite{CHO2013157}. 
The advantage of the single KLE representation is that it requires fewer KL coefficients to represent the correlated fields than the individual KLEs.  
To compute eigenfunctions and eigenvalues in the single KLE representation of $N_p$ parameter fields,  the parameter vectors $\bm{y}_i \in \mathbb{R}^{ N_{xyz}}$ are assembled in a single parameter vector $\bm{y} \in \mathbb{R}^{N_p N_{xyz}}$ as
\begin{equation}\label{eq:assemble_input}
\bm{y} = \left[\bm{y}_1, ..., \bm{y}_{N_p} \right].
\end{equation}
Then, the eigenvalues and eigenvectors in the $\bm{y}$ KLE are found using the SVD decomposition of the data matrix $\bm{X} \in \mathbb{R}^{N_p N_{xyz}\times N_{train}}$ with rows formed by the vectors 
$\bm{y}^{(i)} - \overline{\bm{y}}$, where $\bm{y}^{(i)} \in \mathbb{R}^{N_p N_{xyz}} $ is the parameter vector in the $i$th sample of $D$ and  $\overline{\bm{y}} \in \mathbb{R}^{N_p N_{xyz}} $ is the sample mean of $\{ \bm{y}^{(i)} \}_{i=1}^{N_{train}}$.

\subsection{Evaluation Metrics}
We evaluate the accuracy of the operator learning models and KLE representations using the \ac{MAE}:
\begin{equation}\label{eq:mae}
    MAE = \frac1{N_{xyz}}\frac1{N_t} \frac1{N_{samples}} \sum_{i=1}^{N_{xyz}} \sum_{j=1}^{N_{t}
    } \sum_{k=1}^{N_{samples}}|\mathcal{G}[\bm{y}^{(k)}](\bm{x}_i, t_j) - u^{(k)}(\bm{x}_i, t_j)|
\end{equation}
and root mean square error \ac{RMSE}:
\begin{equation}\label{eq:rmse}
    RMSE = \frac1{N_{samples}} \sum_{k=1}^{N_{samples}} \sqrt{\frac1{N_{xyz}} \frac1{N_t}  \sum_{i=1}^{N_{xyz}} \sum_{j=1}^{N_{t}}
    \left(\mathcal{G}[\bm{y}^{(k)}](\bm{x}_i, t_j) - u^{(k)}(\bm{x}_i, t_j)\right)^2},
\end{equation}
where $N_{samples}$ is the number of samples (usually, the number of test samples) for which the accuracy of the solution is evaluated. 
\ac{RMSE} is more sensitive to large local errors than \ac{MAE}. In this work, we use RMSE to train DNNs in the operator learning model and MAE to optimize hyperparameters, including $N_\eta$, $N_\xi$, and the size of the DNN's hidden layers.

We also use time-dependent $\ell_\infty$ error 
\begin{equation}\label{eq:l_infty_error}
    \ell_\infty (t_j)= \max_{\bm{x}}{\left|\left(\mathcal{G}[\bm{y}^{(k)}](\bm{x}, t_j) - u^{(k)}(\bm{x}, t_j)\right)\right|}
\end{equation}
to study how maximum model errors change over time. 

\section{Numerical Example}\label{sec:ibdp_dataset}

We test the proposed operator learning method for modeling CO$_2$ sequestration at the IBDP site \cite{Finley2014}. 
\ac{IBDP} is a carbon capture and storage pilot project for CO$_2$ injection into a deep saline reservoir.
Over a 3-year period starting in November 2011, approximately 1 million tonnes of CO$_2$ were injected into the reservoir.

\begin{figure}[htbp]
\centering
\includegraphics[width=0.6\textwidth]{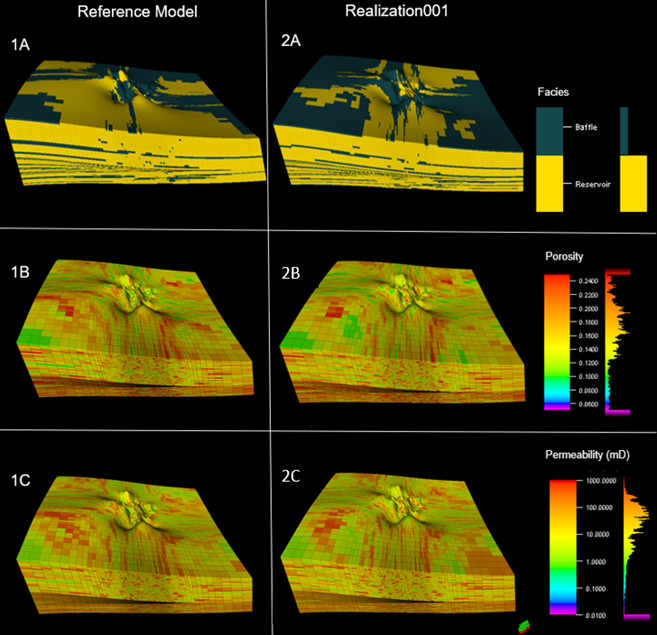}
\caption{Left column: geological facies, porosity, and horizontal permeability from the IBDP calibration study reported in \cite{Finley2014}. Right column: statistical realizations of the geological facies, porosity, and horizontal permeability.}
\label{fig:ibdp_geology}
\end{figure}

To train and test the operator learning model, we use a dataset from \cite{KADEETHUM2024213007} consisting of 100 simulations of the CO$_2$ pressure and saturation on a $(126, 125, 110)$ unstructured grid. The simulations were performed using the ECLIPSE software (\cite{pettersen2006basics}). Estimates of reservoir parameters, including horizontal and vertical permeability ($k_h$ and $k_v$) and porosity ($n$), were statistically generated using specified mean and covariance models. The statistical models were developed as part of the IBDP model parameterization study, using stratigraphic characterization and analysis of seismic data \cite{Finley2014}. The facies, porosity, and horizontal permeability from this study, as well as one realization of these properties using the statistical model, are shown in Figure \ref{fig:ibdp_geology}.

\begin{figure}[htbp]
\centering
\includegraphics[width=0.8\textwidth]{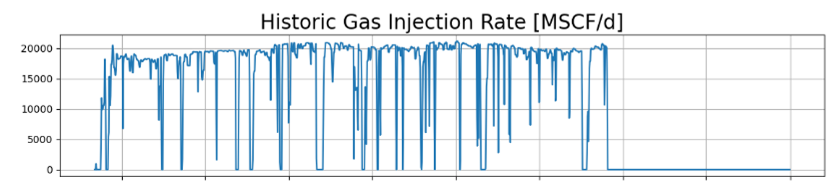}
\caption{The CO$_2$ injection schedule used for generating the training and testing data based on the historic injection rates at the IBDP site.}
\label{fig:injection_rate}
\end{figure}

For dataset generation, the statistical model was further modified to match the observed trend in CO$_2$ saturation at an observation well at the IBDP site. A modifier was applied to the vertical permeability to contain the CO$_2$ plume to the lower part of the reservoir, as was observed in the field.  The resulting dataset contains 100 samples in total, including 80 with modifiers and 20 without modifiers. All solutions were obtained for the same CO$_2$ injection schedule, shown in Figure \ref{fig:injection_rate}. The dataset provides CO$_2$ pressure $p$ and saturation $S$ monthly over 50 months.

In this study, we construct separate operator-learning models for permeability fields with and without modifiers. Ninety percent of the samples are used for training, while ten percent are reserved for testing. 
Furthermore, we train 
$\mathcal{G}_p [y_h,y_v,n](\bm{x}, t)$
and
$\mathcal{G}_S [y_h,y_v,n] (\bm{x}, t)$ models for pressure and saturation, respectively, where $y_h = \ln k_h$ and $y_v = \ln k_v$. The three input variables are represented with a single KLE as described in Section \ref{sec:dynamic_kl_dnn}. In the following, we use superscripts 80 and 20 to denote the operator learning models trained for the permeability fields with and without modifiers, respectively. For example, $\mathcal{G}_p^{80} [y_h,y_v,n](\bm{x}, t)$ is the pressure operator learning model trained on permeability fields with modifiers. 
The representation of inputs in the operator-learning models, including the size of the data matrices, $N_\xi$, time and memory required for constructing the KLEs, and errors in approximating parameters due to KLE truncation, are summarized in Table \ref{table:inputs}. 

\begin{table} 
\begin{center} 
\caption{KLE representation of input fields $y_h$, $y_v$, and $n$ in the IBDP dataset: 
the training dataset size $N_{train}$;
the size of the data matrices for constructing a joint KLE of the input fields ($\bm{X}$ size), 
the number of terms, $N_\xi$, in the joint KLE expansion of the input fields;
time for constructing the joint KLE of the input fields (KLE time);
computer memory required for performing SVD (KLE memory); and 
RMSEs in approximating $y_h$, $y_v$, and $n$ fields with KLEs (KLE RMSE).}
\label{table:inputs}
\begin{tabular}{ |c|c|c|c|c|c|c|c|c| } 

 \hline
  \makecell{Model} & \makecell{$N_{train}$} & \makecell{$\bm{X}$ \\ size} & \makecell{$N_\xi$} & \makecell{\ac{KLE} \\ time} & \makecell{\ac{KLE} \\ memory} & \makecell{\ac{KLE} $y_h$\\ RMSE} & \makecell{\ac{KLE} $y_v$\\ RMSE} & \makecell{\ac{KLE} $n$\\ RMSE} \\ 
 \hline
 $\mathcal{G}_p^{80}$ / $\mathcal{G}_s^{80}$ & 72 & \makecell{(5.2M, 72) }  & 18 & 33 s & 3.2 GB & 1.766 & 1.683 & 0.0319\\ 
 \hline
 $\mathcal{G}_p^{20}$ / $\mathcal{G}_s^{20}$ & 18 & \makecell{(5.2M, 18) }  & 9 & 11 s & 1.3 GB & 1.650 & 1.650 & 0.0322\\
 \hline

\end{tabular}
\end{center}
\end{table}

In the dataset, the CO$_2$ pressure and saturation are computed over the entire computational domain, with a sample size of $N^p_{xyzt}= 126 \times 125 \times 110 \times 50$. We train $\mathcal{G}_p [y_h,y_v,n](\bm{x}, t)$ to model pressure for the entire domain using the full samples of the $p$ solution. On the other hand, when training $\mathcal{G}_S [y_h,y_v,n](\bm{x}, t)$ model, we observe that in all simulations in the dataset, the CO$_2$ plume is confined to a relatively small subdomain ($40 \times 44 \times 94$ grid points) centered on the injection well. 
Therefore, we train $\mathcal{G}_S [y_h,y_v,n](\bm{x}, t)$ for the CO$_2$ saturation modeling only in this subdomain, reducing the sample size of the $S$ solution to $N^S_{xyzt}=40 \times 44 \times 94 \times 50 = 165440 \times 50 = 8.272M$. The resulting sizes of the data matrices $\bm{X}$ used to perform low-rank SVD for representing pressure and saturation in the $\mathcal{G}_p$ and $\mathcal{G}_S$ models, as well as the number of KL terms, time to and memory required for constructing KLEs, the approximation errors, and KL-DNN model errors are listed in Table \ref{table:parameters}.

In the context of $\mathcal{G}_p$ models, it is essential to utilize a low-dimensional Karhunen-Loève expansion (KLE) for the pressure field. This approach is necessary because performing a low-rank singular value decomposition (SVD) on $\bm{X} \in \mathbb{R}^{72 \times 86.6M}$ for the full-dimensional KLE of pressure is computationally infeasible. The full-dimensional KLE would require storing an $\bm{U}$ matrix of size 86.6M × 86.6M, which demands approximately 29,600 petabytes of memory. This amount of memory exceeds the capabilities of one of the largest existing supercomputers, El Capitan, by nearly six times \cite{elcapitan_llnl}.
The $\bm{y}$ parameter in the pressure model and both $\bm{y}$ and $S$ in the saturation model are modeled with full-dimensional KLEs, as the sizes of the data matrices for these variables are small enough for performing low-rank SVDs. 

The DNN models in $\mathcal{G}_p [y_h,y_v,n](\bm{x}, t)$
and
$\mathcal{G}_S [y_h,y_v,n] (\bm{x}, t)$ have 3750 and 1957 trainable parameters, respectively. Both DNN models are trained on an NVIDIA RTX A6000 GPU; the total training time of all four models ($\mathcal{G}_p^{80}$, $\mathcal{G}_p^{20}$, $\mathcal{G}_S^{80}$, and $\mathcal{G}_S^{20}$) is around 20 minutes, and inference time is less than 1 minute per model. 

First, we study the accuracy of the low-dimensional KLE approximation relative to the full dimensional approximation.
Since the full-dimensional KLE of $p$ cannot be constructed, we compare the KLE approximation RMSEs in the full- and low-dimensional KLEs of the saturation field defined on the smaller domain centered around the injection well. An approximation RMSE is defined according to Eq \eqref{eq:rmse} with $\bm\eta$ in $\mathcal{G}[\bm{y}^{(j)}]$ determined from the inverse KLE operator \eqref{eq:inverseKL}.   
We find that the low-dimensional KLE increases the approximation (\ac{RMSE}) by less than 2\% as compared to the full-dimensional KLE.   
This comparison demonstrates that a low-dimensional KLE can be used to construct an accurate low-dimensional representation of massive datasets without resorting to spatial coarsening or domain subsampling. 

\begin{table} 
\begin{center} 
\caption{Performance and characteristics of the KL-DNN models used in this study: 
the training dataset size $N_{train}$;
the testing dataset size $N_{test}$;
$\bm{X}$ size is the size of the data matrices for constructing a full or nested KLE of the state variables ($p$ or $S$) input fields;
the number of terms ($N_\eta$ and $N_\gamma$) in the low-dimensional KLE expansion of pressure and saturation ($N_\gamma$ is not listed for $\mathcal{G}_S$ models because the full-dimensional KLEs of $S$ are used);
KLE time is the total time for training the operator learning models (the DNN training in all models takes about 5 minutes);
KLE memory is the computer memory required for performing SVD of the $p$ and $S$ data matrices; 
KLE RMSE is the RMSE in approximating $p$ and $S$ fields with KLEs;
and KL-DNN RMSE is the total RMSE in the operator-learning model predictions.}
\label{table:parameters}
\begin{tabular}{ |c|c|c|c|c|c|c|c|c|c| } 

 \hline
  \makecell{Model} & \makecell{$N_{train}$} & $N_{test}$ &  \makecell{$\bm{X}$ \\ size} & \makecell{$N_\eta$} & \makecell{$N_\gamma$} & \makecell{\ac{KLE} \\ time} & \makecell{\ac{KLE} \\ memory} & \makecell{\ac{KLE} \\ RMSE} & \makecell{\ac{KL-DNN} \\ RMSE} \\ 
 \hline
 $\mathcal{G}_p^{80}$ & 72 & 8 & \makecell{(86.6M, 72) }  & 4 & 3 & 1000 s & 1 GB & 1.045 psi & 1.139 psi\\ 
 \hline
 $\mathcal{G}_p^{20}$ & 18 & 2 & \makecell{(86.6M, 18) }  & 3 & 2 & 150 s & 0.4 GB & 0.811 psi & 0.956 psi\\
 \hline
$\mathcal{G}_S^{80}$ & 72 & 8 & (8.3M, 72)  & 6 & - & 50 s & 2.3 GB & 0.0140 & 0.0148\\
\hline
 $\mathcal{G}_S^{20}$ & 18 & 2 & (8.3M, 18)  & 6 & - & 15 s & 1.5 GB & 0.0134 & 0.0138\\
  \hline

\end{tabular}
\end{center}
\end{table}

Next, we discuss the performance and accuracy of the KL-DNN models for the pressure and saturation fields as summarized in Table \ref{table:parameters}.  
The reported training time, memory, approximation errors, and model errors are all averages over test cases. The number of test samples for each operator learning model is given in the ``$N_{test}$'' column.  
The training times reported in the ``\ac{KLE} time'' column include time spent computing eigenfunctions for both state variables and parameters, as well as for computing the reduced (latent space) dataset $D_l$. 
In the $\mathcal{G}_p^{80}$ and $\mathcal{G}_p^{20}$ pressure models, the training times are higher because low-dimensional SVDs are performed for each of 50 time steps. The KL-DNN model RMSEs reported in the ``KL-DNN RMSE'' column are a combination of the KL approximation RMSE and the DNN mapping error. The approximation and operator learning RMSEs in  $\mathcal{G}_p^{80}$ and $\mathcal{G}_S^{80}$ models are larger than in the  $\mathcal{G}_p^{20}$ and $\mathcal{G}_S^{20}$ models despite the fact that the training dataset for $\mathcal{G}_p^{80}$ and $\mathcal{G}_S^{80}$ models is four times larger. 
This is because the pressure and saturation fields in the model without modifiers are smoother and can be represented more accurately with the same or fewer latent variables than those in the model with modifiers as evident by the KLE approximation errors in the ``KLE RMSE'' column.

The RMSEs reported in Table \ref{table:parameters} are obtained by replacing $\bm{W}$ with the identity matrix as in the standard PCA expansions, because we find that the resulting KLEs have lower approximation errors.

\begin{figure}[htbp] 
    \centering
    \includegraphics[width=.6\textwidth]{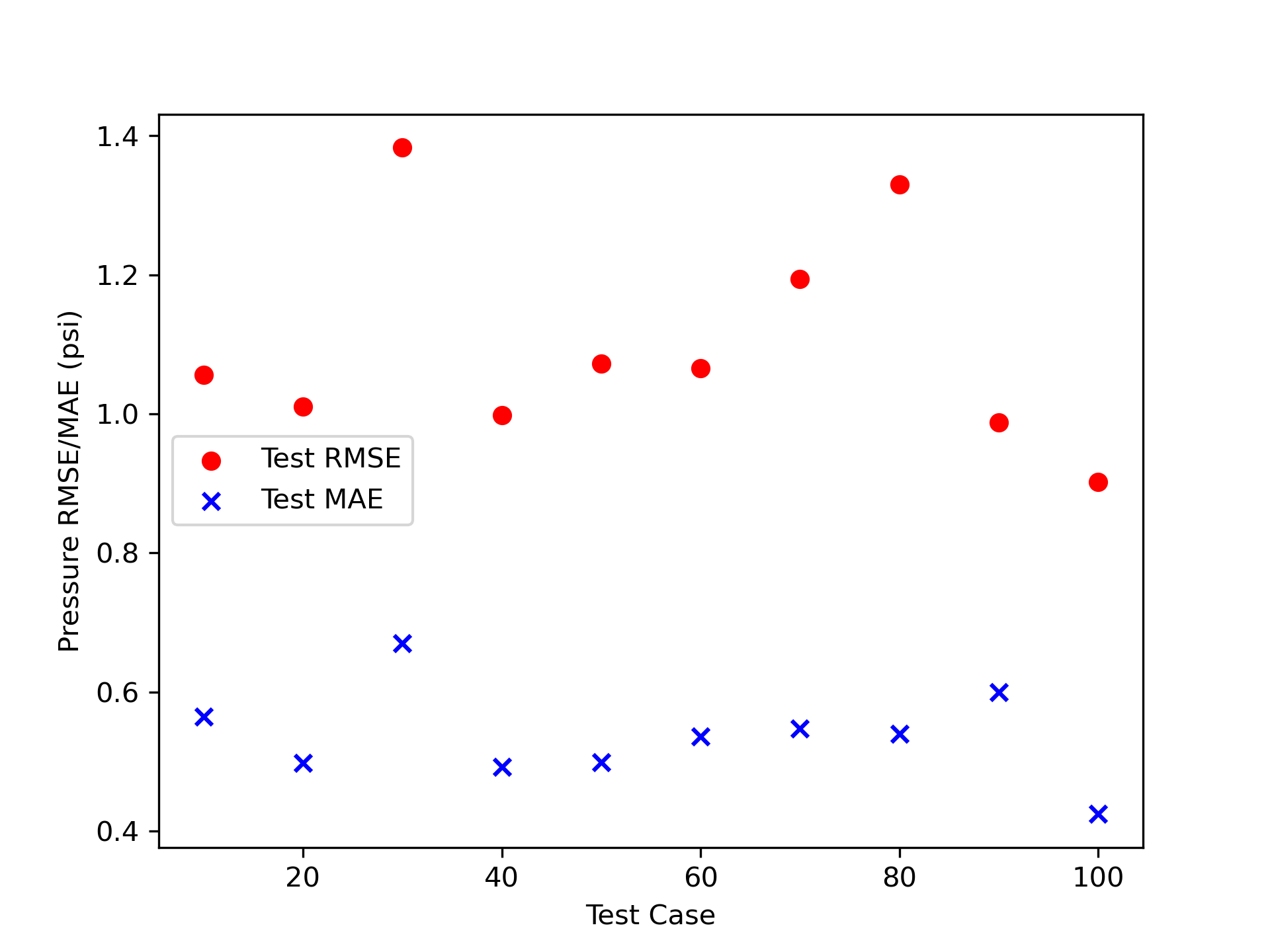}
    \caption{RMSE and MAE of the fluid pressure predictions for 10 test cases}
    \label{fig:pressure_rmse}
\end{figure}

Figure \ref{fig:pressure_rmse} reports \ac{RMSE} and MAE values in the pressure predictions for the 10 test cases, where the first eight cases are used for testing $\mathcal{G}_p^{80}$ and the last two cases are for testing $\mathcal{G}_p^{20}$. 
  
RMSEs range between 0.9 psi and 1.4 psi across all test cases, with the average RMSE over the 10 test cases of 1.1 psi. The average pressure in the simulations is 3150 psi, yielding the relative average RMSE of 0.04\%.

\begin{figure}[htbp] 
    \centering
    \includegraphics[width=1.0\textwidth]{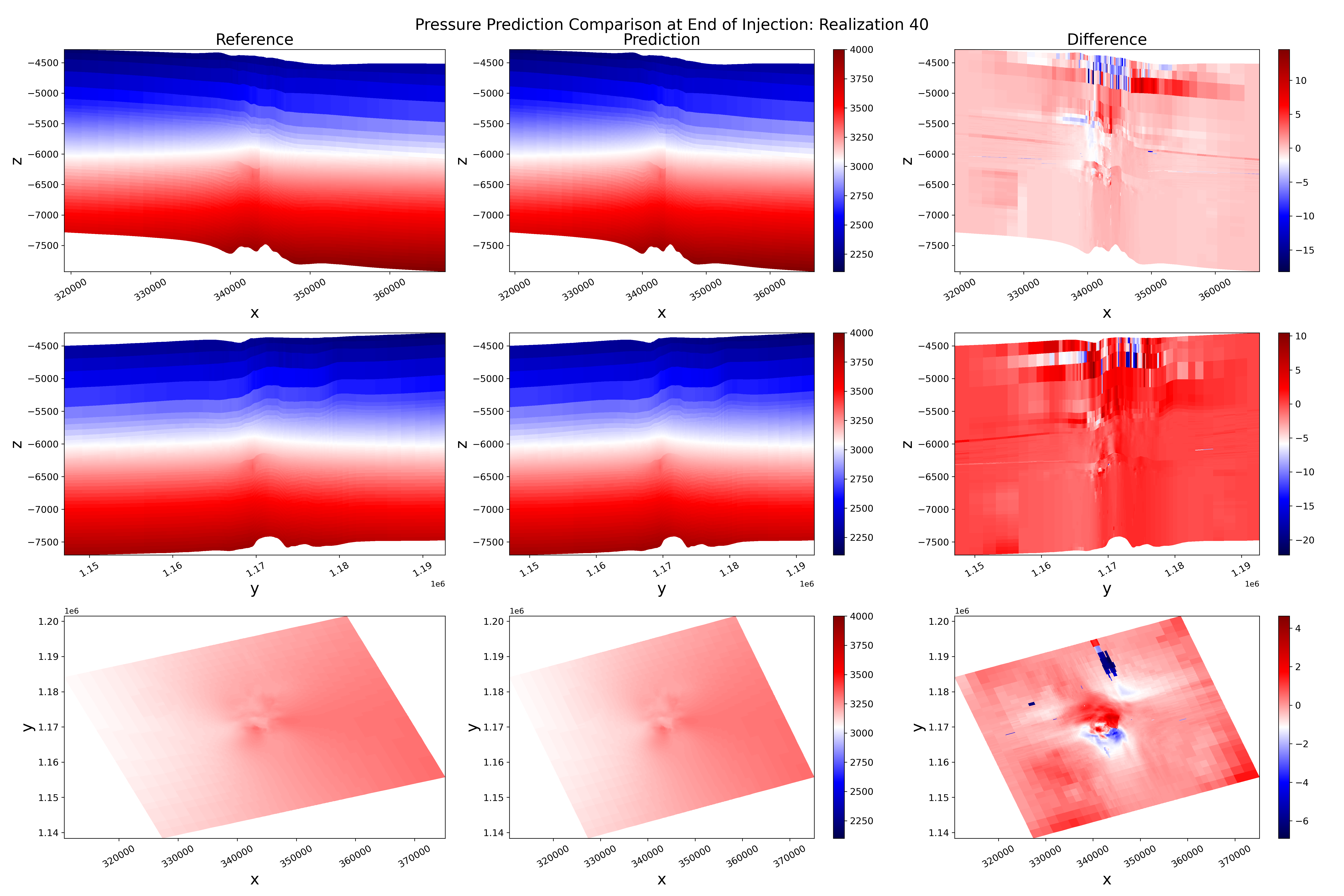}
    \caption{Pressure reference value, prediction, and point error slices across the injection well for the test case 40 at the end of injection (36 months). The model is trained on the dataset with permeability modifiers.}
    \label{fig:pressure_40}
\end{figure}

Figure \ref{fig:pressure_40} shows cross-sections of reference and predicted pressure fields as well as point errors in the predicted pressure field at the end of injection (36 months) for one of the test cases.
The pressure field is relatively smooth, except for a small area near the injection well, where we observe the largest point errors. In some test cases, we also observe elevated point errors close to the boundaries of the spatial domain.

\begin{figure}[htbp] 
    \centering
    \includegraphics[width=0.6\textwidth]{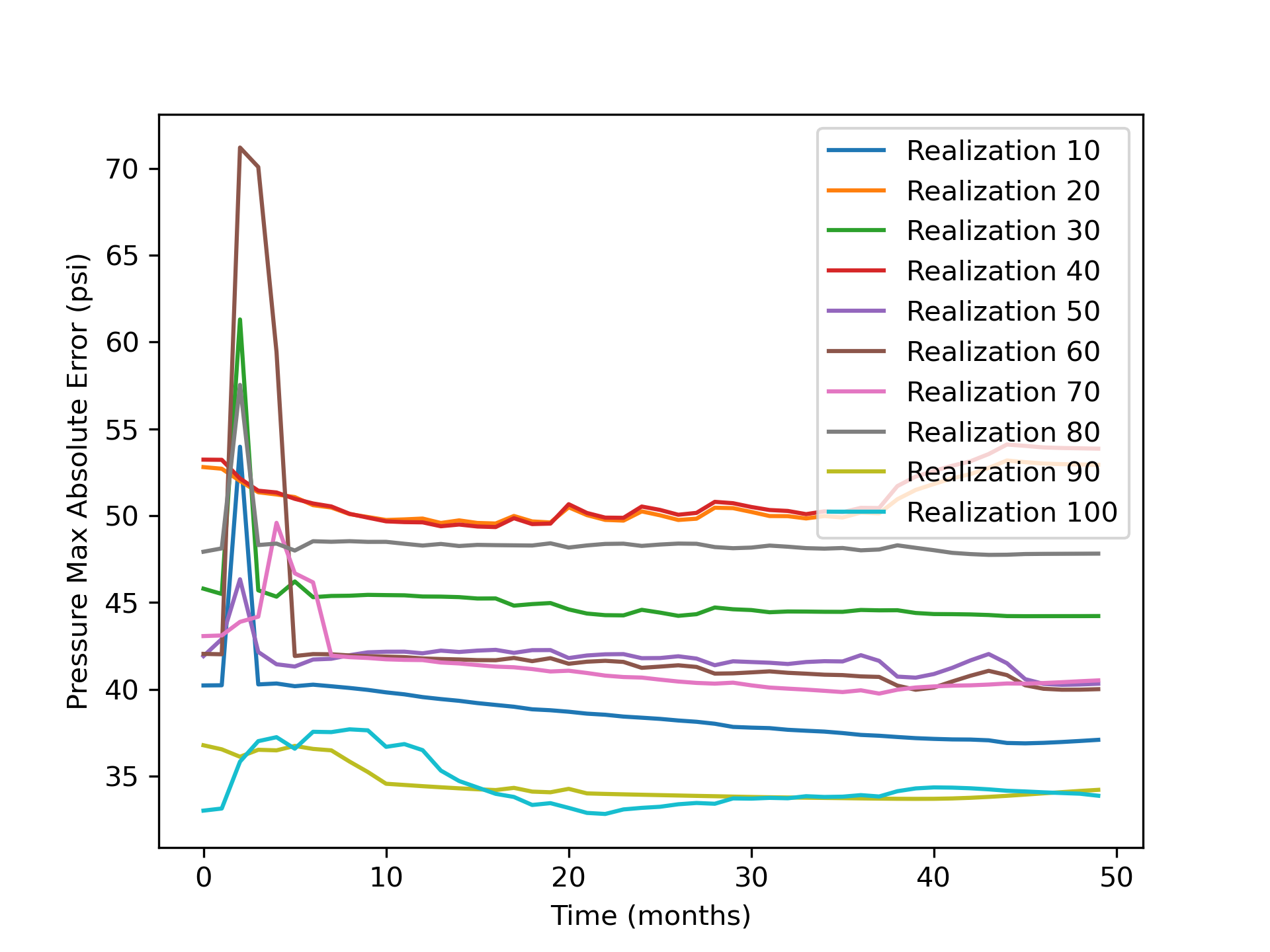}
    \caption{Maximum Absolute Error as a function of time of the fluid pressure predictions for 10 test cases.}
    \label{fig:pressure_mae_time}
\end{figure}

Figure \ref{fig:pressure_mae_time} depicts $\ell_\infty$ errors, the maximum absolute errors, in pressure predictions as functions of time for the 10 test cases. The errors reach their maximum values during the first five months after injection begins, when the largest pressure changes occur. 
The fact that the errors decrease after the initial injection phase and do not increase further over time indicates that the model provides relatively accurate long-term pressure predictions. 

\begin{figure}[htbp] 
    \centering
    \includegraphics[width=0.6\textwidth]{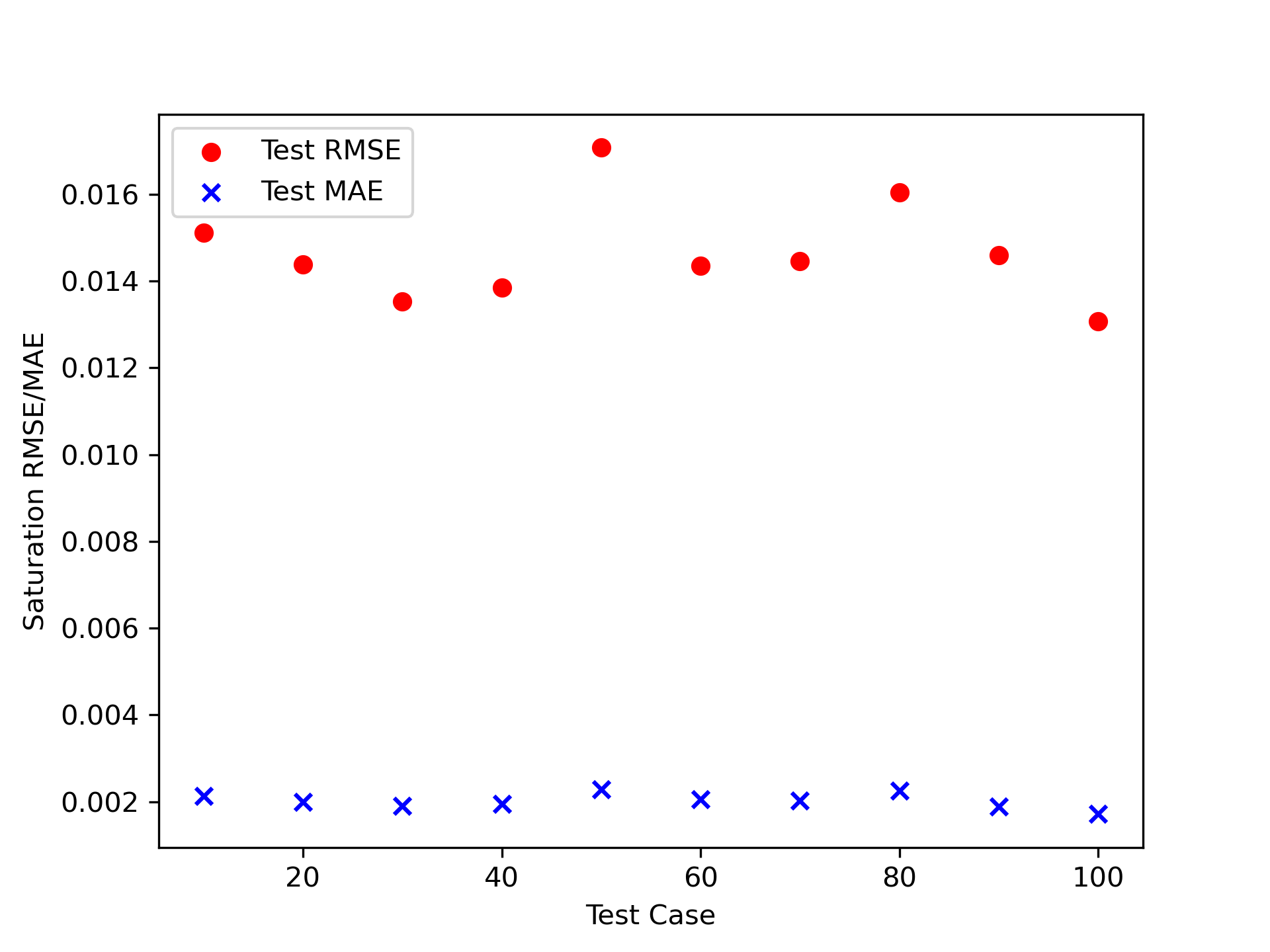}
    \caption{RMSE and MAE of saturation predictions for 10 test cases.}
    \label{fig:saturation_rmse}
\end{figure}

\begin{figure}[htbp] 
    \centering
    \includegraphics[width=0.8\textwidth]{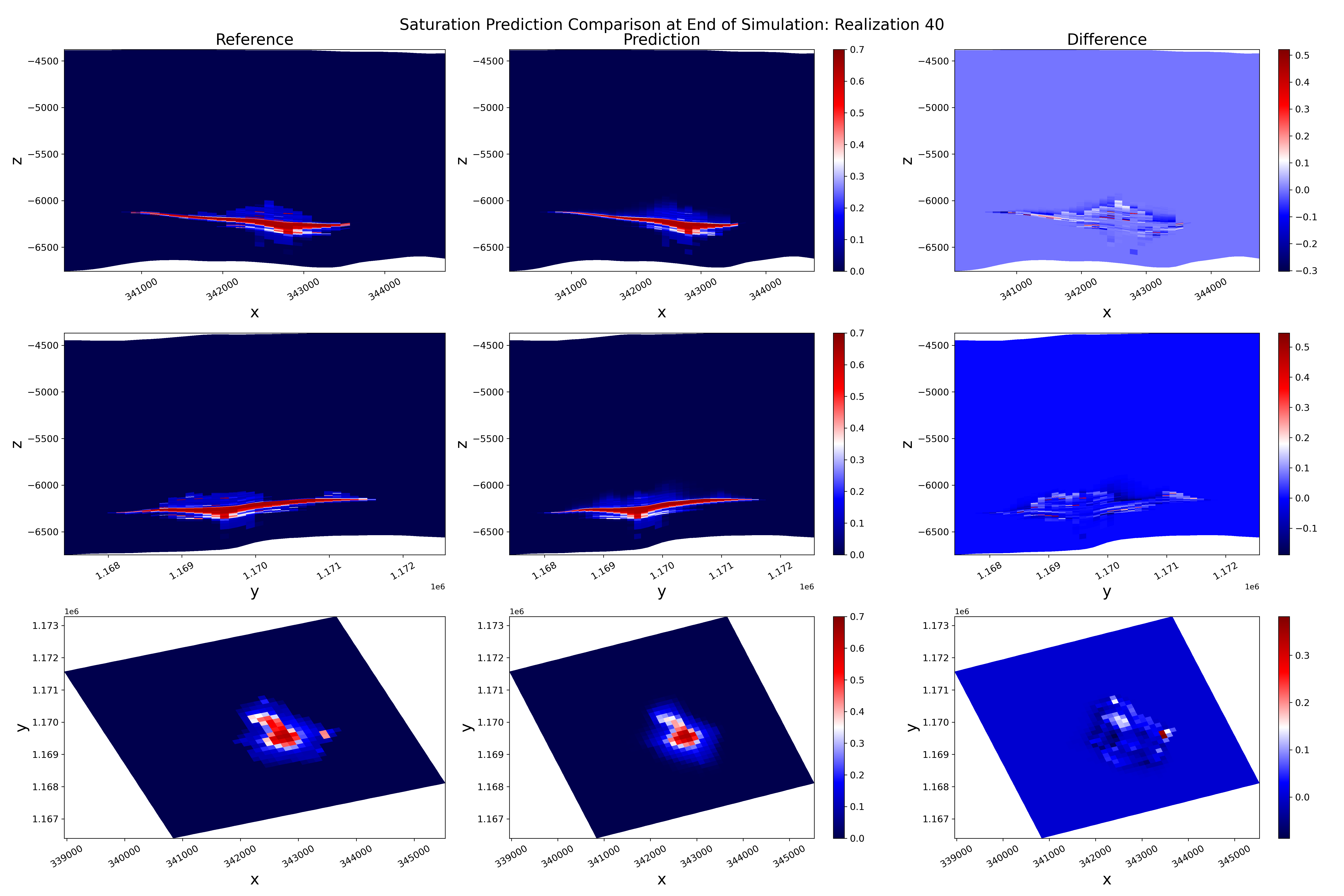}
    \caption{$CO_2$ saturation: cross-sections showing reference values, predictions, and point errors for test case 40. The model is trained on the dataset with permeability modifiers.}
    \label{fig:saturation_70}
\end{figure}

\begin{figure}[htbp] 
    \centering
    \includegraphics[width=0.6\textwidth]{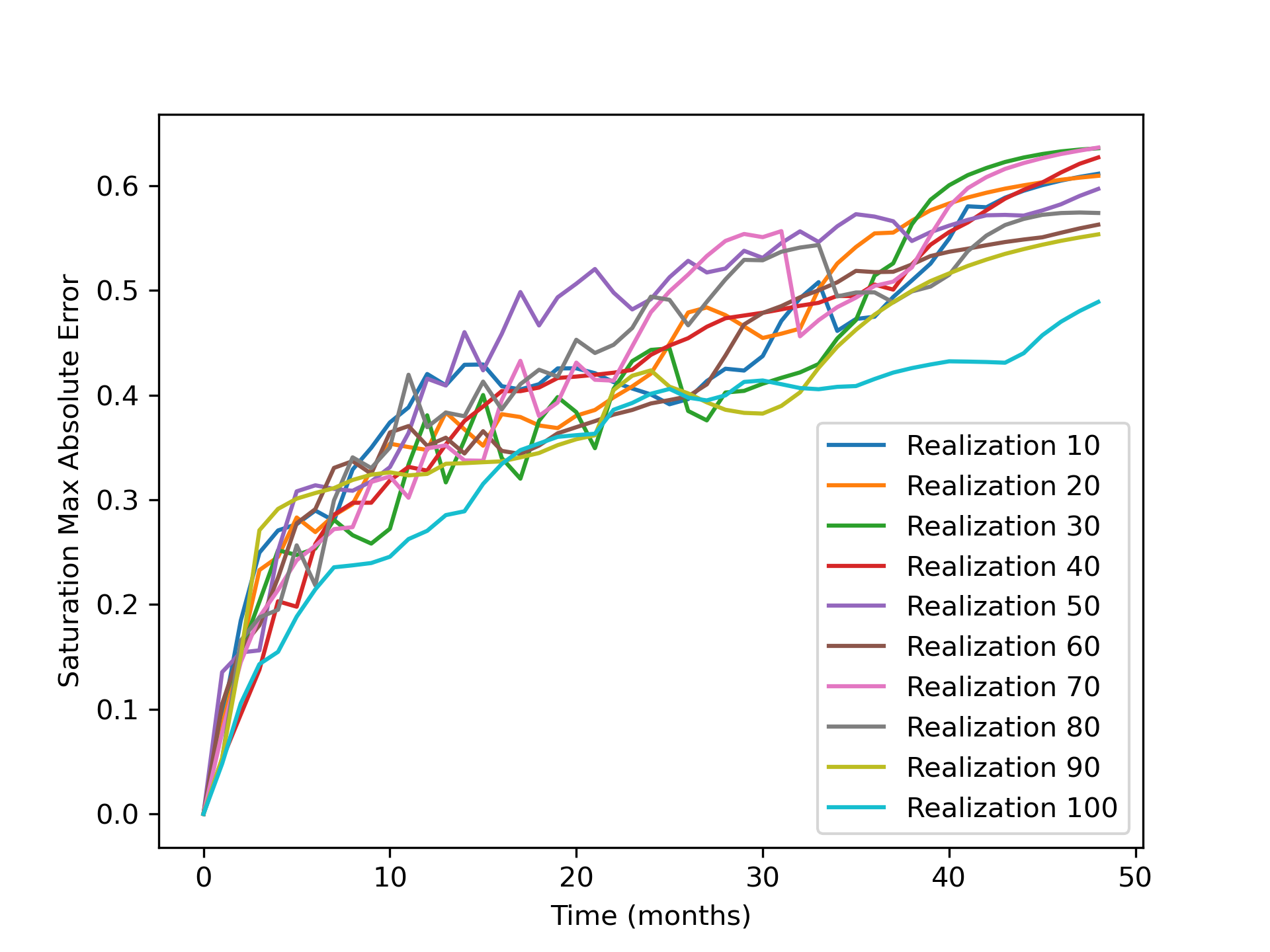}
    \caption{Maximum absolute error in saturation predictions as functions of time for 10 test cases. }
    \label{fig:saturation_mae_time}
\end{figure}

Figure \ref{fig:saturation_rmse} displays RMSEs and MAEs in the saturation predictions for the 10 test cases. We find that the errors are similar across all test cases, with an average RMSE of 0.015 and an average MAE of 0.002. 
Figure \ref{fig:saturation_70} shows the reference and predicted CO$_2$ saturation fields for one of the test problems. The largest point errors are observed at the CO$_2$ plume boundaries.  
Figure \ref{fig:saturation_mae_time} displays $\ell_\infty$ errors in the saturation predictions as functions of time for the 10 test cases. We find that the errors increase with time. 
Nonetheless, the operator learning model preserves the overall plume shape and provides accurate saturation predictions within the plume.

\begin{figure}[htbp] 
    \centering
    \includegraphics[width=0.6\textwidth]{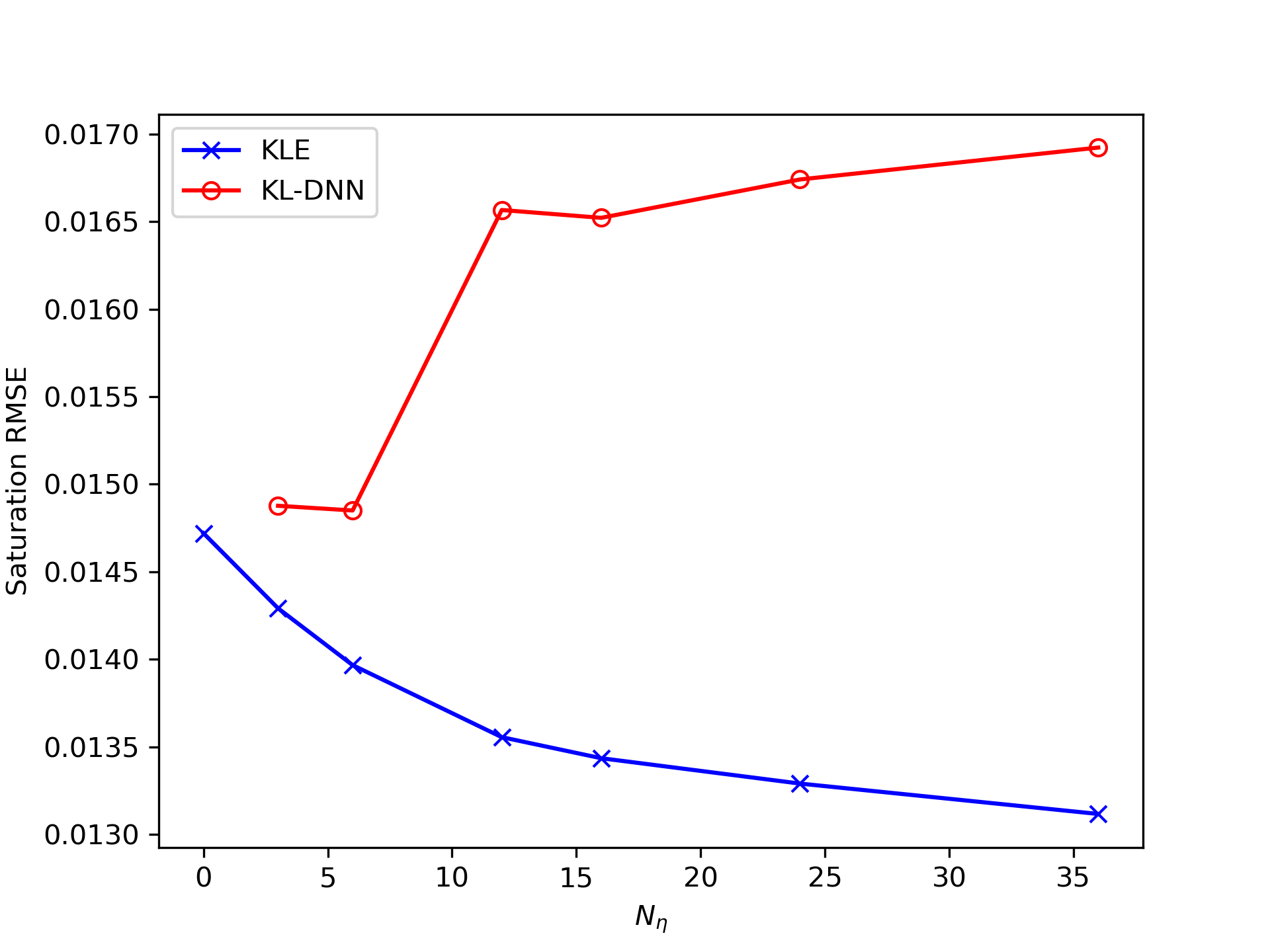}
    \caption{KLE RMSE (KLE approximation error) and KL-DNN RMSE (KL-DNN model error) as functions of $N_\eta$. RMSEs are averaged over the 10 test cases.}
    \label{fig:saturation_errors}
\end{figure}

The average RMSE in the KL-DNN $S$ predictions over all 10 test cases is 0.0146 or 5\% relative to the average saturation in the plume, which is larger than the error in the pressure prediction (0.04 \%). 
The average RMSE and relative errors in KL-DNN $p$ and $S$ models are summarized in Table \ref{table:comparison}. For comparison, Table \ref{table:comparison} also lists average RMSEs in the DeepONet model predictions as reported in \cite{KADEETHUM2024213007}. The DeepONet model was trained on the same training dataset and evaluated on the same testing dataset as the KL-DNN model in this study.  We note that the KL-DNN RMSE errors reported in Table \ref{table:comparison} are different from those in Table \ref{table:parameters}, which were computed separately for the input parameters with and without modifiers. 
We find that RMSEs in KL-DNN are lower than those in DeepONet for both pressure and saturation. The average saturation RMSE in our model is 0.015, which is slightly smaller than the RMSE of 0.017 reported in \cite{KADEETHUM2024213007}. The average pressure RMSE of 1.1 psi in our model is approximately 20\% smaller than that in \cite{KADEETHUM2024213007}.  Moreover, the training time of the DeepONet model in \cite{KADEETHUM2024213007} (2700 minutes) is about 2 orders of magnitude larger than that of our model. We note that the DeepONet model in \cite{KADEETHUM2024213007} employed subsampling for training on the IBDP dataset.

\begin{table} 
\begin{center} 
\caption{RMSE and training time of the KL-DNN and DeepONet pressure and saturation predictions averaged over 10 test cases. The DeepONet results are reported based on  \cite{KADEETHUM2024213007}.}
\label{table:comparison}
\begin{tabular}{ |c|c|c| } 

 \hline
  \makecell{Model} &  Training time (min) & Testing Cases RMSE  \\ 
 \hline
 $\mathcal{G}_p$ & 20 & 1.10 psi \\ 
 \hline
 
$\mathcal{G}_S$ & 6 & 0.0146 \\
\hline
 Pressure DeepONet \cite{KADEETHUM2024213007} & 2700 & 1.35 psi\\ 
 \hline
 Saturation DeepONet \cite{KADEETHUM2024213007} & 2700 & 0.0157 \\
  \hline

\end{tabular}
\end{center}
\end{table}

Finally, we comment on the approximation and model RMSEs in pressure and saturation reported in Table \ref{table:parameters}. As noted earlier, the model RMSE is a combination of the KLE approximation error and the error due to mapping of $\bm\xi$ to $\bm\eta$. In Table \ref{table:parameters}, the model RMSEs are dominated by the KLE approximation errors. The approximation errors can be reduced by including more terms in the KLEs of $p$ and $S$.   
In Figure \ref{fig:saturation_errors}, we demonstrate that the RMSE in the KLE approximation of $S$, averaged over the 10 test cases, decreases as a function of $N_\eta$, the number of terms in the KLE expansion of the saturation field. Figure \ref{fig:saturation_errors} also shows the average RMSE of the KL-DNN predictions of $S$, which reaches a minimum for $N_\eta=6$. As we can see, for $N_\eta > 6$, the approximation error decreases, but the model error increases with increasing $N_\eta$. This is because for $N_\eta >6$, there are not enough samples in the training dataset to train the DNN, whose size increases with $N_\eta$.

\section{Conclusions}\label{sec:conclusion}
We proposed a scalable operator learning model and applied it to the large-scale IBDP \ac{GCS} problem. Our model offers faster training and higher accuracy than the DeepONet model trained on the same dataset.

Like the DeepONet operator-learning model, the proposed model performs operator learning in the latent spaces of the system's states and parameters. Unlike the DeepONet, our model trains the parameter encoder (equivalent to the leading part of the DeepONet branch), the DNN map (equivalent to the back end of the DeepONet branch), and the decoder (equivalent to the trunk of the DeepONet branch) separately. This trainable-by-part feature of our model, in combination with a low-dimensional KL representation of high-dimensional states and low-rank SVD, enables it to scale to large-dimensional problems. DeepONet and other operator-learning models often use subsampling for large-scale problems, which can introduce additional errors. 

The training time for our model, which includes both SVD and DNN training, is approximately 20 minutes. This is two orders of magnitude less than the training time required for the DeepONet model.
Our model has average pressure and saturation RMSEs of 1.1 psi and 0.0146, respectively, which are lower than the DeepONet model's errors (1.35 psi and 0.0157, respectively). 
Point errors in the pressure predictions are higher near the injection well and the domain boundaries. Saturation point errors are highest near the edge of the plume.

Our results demonstrate the utility of the operator learning models for practical large-scale applications such as GCS. The inference time of less than 1 minute per simulation makes it well-suited for uncertainty quantification, history matching, and other tasks that require multiple model evaluations. The training time of less than 20 minutes on a regular workstation makes the model accessible to most users.

The proposed trainable-by-part approach and low-dimensional KL representation can be used for scaling other operator learning models, including PCA-Net and DeepONet, to large-scale problems.

\section*{Conflict of Interest disclosure}
The authors declare there are no conflicts of interest for this manuscript.

\section*{Acknowledgments}

This research was partially supported by the Strategic Research Initiative (SRI) Program at the  Grainger College of Engineering of the University of Illinois Urbana - Champaign.

\bibliographystyle{elsarticle-num}
\bibliography{sample}

\end{document}